\ifx\XeTeXversion\undefined\pdfoutput=1\fi
\documentclass[11pt]{article}
\usepackage[utf8]{inputenc}
\usepackage[T1]{fontenc}
\usepackage{lmodern}
\usepackage{microtype}
\usepackage[margin=1in]{geometry}
\usepackage{amsmath,amssymb,amsfonts}
\usepackage{booktabs}
\usepackage{multirow}
\usepackage{graphicx}
\usepackage{xcolor}
\usepackage{tikz}
\usetikzlibrary{arrows.meta,positioning,fit,calc,backgrounds}
\usepackage{pgfplots}
\pgfplotsset{compat=1.18}
\usepackage[hypertexnames=false,colorlinks=true,linkcolor=black,citecolor=black,urlcolor=blue!60!black]{hyperref}
\usepackage{enumitem}

\definecolor{paceblue}{RGB}{0,102,166}
\definecolor{retyellow}{RGB}{204,145,0}
\definecolor{h2ogray}{RGB}{120,120,120}
\definecolor{boxblue}{RGB}{224,240,250}
\definecolor{boxgray}{RGB}{242,242,242}

\title{\textbf{PACE: Keeping the Right Tokens in the KV Cache\\Is Not Enough for Long-Horizon LLM Agents}}
\author{Suman Chatterjee\\Amazon Web Services}
\date{}

\begin{document}
\maketitle

\begin{abstract}
\noindent Long-horizon LLM agents accumulate sessions of up to millions of tokens, far more key-value (KV) cache than one GPU holds, so serving systems keep a small budget of tokens and evict the rest. Eviction methods differ in which tokens they keep but share one goal: that the tokens a later step needs are still cached. We show this is not enough. In 100K-token coding sessions on Qwen3-8B with a 2,048-token budget, a planner-guided policy that keeps over $99\%$ of the needed tokens resident still fails late steps increasingly often: its accuracy falls from $93\%$ to $64\%$ as the gap between a definition and its use grows from 2K to 90K tokens, at residency $0.996$. The needed tokens are \emph{resident but unretrievable}: what grows is their rotary-position distance to the query. We present PACE (Planner-Aware Context Engine), a KV-cache engine that couples plan-aware retention, in which the agent's planner tells the cache which context future steps will read, with bounded positions, which renumber kept tokens so their rotary distance to the query stays below 8K at any session length. In a pre-registered evaluation on fresh tasks, PACE passes $97\%$ of hidden unit tests at 100K tokens, against $42\%$ for $\text{H}_2\text{O}$ and $3\%$ for SnapKV under the same budget, and needs both components: either alone reaches $63\%$ (the combination beats each, $p < 10^{-4}$). The result holds with length: in 1M- and 2M-token sessions on one 24\,GB GPU, with definitions written 0.9M and 1.8M tokens before use, PACE passes $99\%$ and $100\%$ of checks and recalls every session fact, while bounded positions without foresight falls from $33\%$ to $13\%$; PACE holds under $0.7\text{\,GB}$ of KV state where the full cache would need $147\text{--}295\text{\,GB}$. Renumbering is exact, costs no measurable perplexity, and adds $9\%$ wall time. On Mistral-7B, whose coding ability caps it at $40\%$ even with nothing evicted, PACE stays at that ceiling while retention alone falls below it.
\end{abstract}

\section{Introduction}
LLM agents that write code, operate tools, or carry out multi-step plans do not answer one prompt; they run a session. Every step appends an instruction, retrieved material, and the model's own output, and later steps depend on definitions made much earlier: a data model declared at step 1, a business rule fixed at step 3, an interface agreed at step 10. Sessions of $10^5\text{--}10^6$ tokens are routine for repository-scale coding or long tool-use traces. The key-value (KV) cache that attention needs grows linearly with the session: for an 8B-parameter model with grouped-query attention (Qwen3-8B), it costs 144\,KiB per token, so 14.7\,GB at 100K tokens and 147\,GB at 1M. A single 24\,GB accelerator cannot hold it, so serving systems fix a KV budget and evict.

A large body of work studies which tokens to keep. Attention-based policies retain ``heavy hitters'' that past queries attended to \cite{liu2023scissorhands,zhang2023h2o}, query-aware policies score tokens against the current prompt \cite{cai2024pyramidkv,li2024snapkv,tang2024quest}, and streaming policies keep attention sinks plus a recent window \cite{xiao2024streamingllm}. More recently, agent-aware systems exploit knowledge of the workflow to predict which context will be reused \cite{matam2026memdecay,pan2025kvflow,wang2026actkv,zhang2026cachescout,zheng2026pbkv}. These methods differ in how they choose, but they share one objective: when a later step needs a token, the token should still be in the cache.

\begin{figure}[t]
\centering
\begin{tikzpicture}
\begin{axis}[
    width=11.5cm,
    height=4.8cm,
    ymin=0, ymax=106,
    symbolic x coords={2K, 10K, 25K, 50K, 90K},
    xtick=data,
    ytick={0, 25, 50, 75, 100},
    xlabel={Gap between definitions and their use (logical tokens)},
    ylabel={Hidden-test checks passed (\%)},
    ymajorgrids=true,
    grid style={dotted, gray!60},
    legend style={at={(1.03,0.5)}, anchor=west, draw=none, font=\small},
    tick label style={font=\small},
    label style={font=\small}
]
\addplot[color=paceblue, mark=*, line width=1.2pt, mark size=2.2pt] coordinates {
    (2K, 97) (10K, 97) (25K, 100) (50K, 100) (90K, 100)
};
\addlegendentry{PACE ($L=8\text{K}$)}

\addplot[color=retyellow, mark=square*, line width=1.2pt, mark size=2.2pt] coordinates {
    (2K, 93) (10K, 76) (25K, 76) (50K, 77) (90K, 64)
};
\addlegendentry{retention only}

\addplot[color=h2ogray, mark=triangle*, line width=1.1pt, mark size=2.4pt] coordinates {
    (2K, 44) (10K, 55) (25K, 45) (50K, 53) (90K, 52)
};
\addlegendentry{$\text{H}_2\text{O}$}

\node[anchor=west, text=retyellow!90!black, font=\footnotesize, align=left] at (axis cs:2K, 17) {retention only: needed tokens resident 0.996 at every gap};
\end{axis}
\end{tikzpicture}
\caption{Keeping the right tokens is not enough. 100K-token coding sessions, Qwen3-8B, KV budget 2,048 tokens. Planner-guided retention keeps $99.6\%$ of the tokens each late step needs, yet its accuracy falls as those tokens recede; $\text{H}_2\text{O}$ loses them outright. PACE adds bounded positions, holding the rotary distance to the query below 8K, and stays at $97\text{--}100\%$. Five randomized sessions per point, three probes each.}
\label{fig:fig1}
\end{figure}
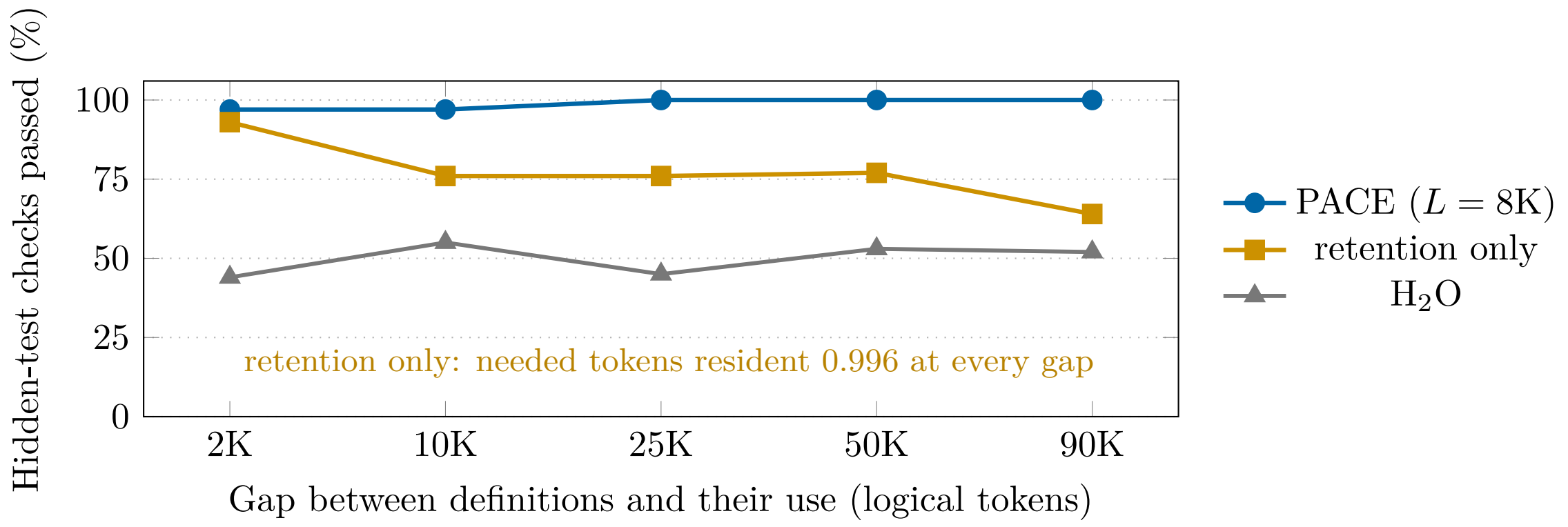

\textbf{Keeping is not enough.} We find that this objective, even when met perfectly, does not deliver correct behavior in long sessions. Figure~\ref{fig:fig1} shows coding sessions of 100K tokens on Qwen3-8B with a budget of 2,048 tokens ($2\%$ of the session). A retention policy guided by the agent's own plan keeps $99.6\%$ of the needed tokens resident at every gap we test, yet its accuracy falls from $93\%$ to $64\%$ as the definitions recede from 2K to 90K tokens before their use. Pooled across configurations, accuracy at full residency declines from $96\text{--}100\%$ when the needed context is within 8K rotary positions of the query to $69\text{--}77\%$ beyond 16K (Figure~\ref{fig:fig3}). The needed tokens are resident but unretrievable: retention fixed which tokens survive, but not where they sit relative to the query in rotary-position space. Standard eviction suffers both problems at once: under the same budget, $\text{H}_2\text{O}$ reaches $40\text{--}55\%$ across our runs and SnapKV $3\%$.

\textbf{PACE.} We present PACE (Planner-Aware Context Engine), a KV-cache engine that addresses both. Plan-aware retention turns the dependency graph an agent planner already produces into a per-span foresight bonus that dominates attention scores, so the context future steps will read is kept, soonest-needed first. Bounded positions renumber the kept tokens in order whenever positions approach a limit $L$, re-rotating their keys exactly, so the rotary distance between any kept token and the query never exceeds $L$, at any session length. With $L = 8{,}192$, the smallest limit the budget allows, needed context stays within 2--7K positions of the query whether it was written 2K or 1.8M tokens earlier. PACE also makes sessions persistent, so each step sends only its new tokens, and spills evicted context to host memory, local storage, and a remote cache from which the planner can restore it.

\textbf{Results.} In a pre-registered evaluation on tasks the method had never seen, PACE passes $97\%$ of hidden unit tests at 100K tokens, against $42\%$ for $\text{H}_2\text{O}$; it recalls every session fact; and it beats each of its two components alone ($p < 10^{-4}$), each of which reaches only $63\%$. The distance sweep isolates the cause: retention alone decays with distance at full residency, while PACE stays at $97\text{--}100\%$ from 2K to 90K. With a production-style LLM planner instead of the true dependency graph, PACE still reaches $92\%$ against $41\%$. The result holds with length: in 1M-token sessions on one 24\,GB GPU, with definitions written about 900K tokens before their use, PACE passes $99\%$ of checks ($74/75$) and recalls every fact ($125/125$), against $33\%$ without foresight and $0\%$ for a StreamingLLM-style cache, with under $0.7\text{\,GB}$ of KV state where the full cache would need $147\text{\,GB}$. At 2M tokens, with definitions 1.8M tokens back, PACE passes all 75 checks and recalls all 125 facts, while bounded positions without foresight falls further, to $13\%$ of checks and $50\%$ recall: the margin from foresight widens with length. On Mistral-7B, whose coding ability caps it at $40\%$ of checks even with nothing evicted, PACE stays at that ceiling from 2K to 90K and recalls $96\text{--}100\%$ of facts, while retention alone drops below it as the gap grows and $\text{H}_2\text{O}$ collapses to $1\%$. Renumbering is token-identical to the baseline when inactive, costs no measurable perplexity, and adds $9\%$ wall time.

\textbf{Scope.} PACE is not a million-token context window. At every step the model attends to about 2K kept tokens, chosen by the plan and by attention; the rest of the session is evicted to host memory and storage. PACE targets agent workloads whose planner can state what later steps depend on, such as coding, multi-step tool use, data pipelines, and multi-agent hand-offs. Without a plan it reduces to attention-based eviction with bounded positions, and full-context serving remains preferable for unplanned recall of arbitrary details from an entire history.

\textbf{Contributions.}
\begin{enumerate}[leftmargin=1.8em, itemsep=2pt, topsep=3pt]
    \item A measured finding: in long-horizon agent sessions under a fixed KV budget, keeping the needed tokens is necessary but not sufficient. At residency above $0.99$, accuracy falls with the rotary-position distance between the needed context and the query (Section~\ref{sec:mechanism}).
    \item PACE, which couples plan-aware retention with bounded, exactly re-rotated positions, plus persistent sessions and tiered rehydration, implemented in an open-source inference engine whose upstream now includes PACE's retention and bounded positions \cite{sparseengine2026} (Section~\ref{sec:pace}). We identify and fix a compounding-rounding failure of naive re-rotation that appears only in very long sessions (Section~\ref{sec:bounded}, Appendix~\ref{app:drift}).
    \item An execution-scored, pre-registered long-horizon benchmark protocol with randomized specifications, leak checks, a no-eviction control, and a recall metric that separates forgetting from coding errors (Section~\ref{sec:setup}), and an evaluation showing at 100K tokens that both components are required and, up to 2M tokens, that PACE keeps its quality (Section~\ref{sec:results}).
\end{enumerate}

\section{Background and Problem}
\label{sec:background}

\textbf{KV eviction under a budget.} An autoregressive transformer caches, for every past token and every layer, a key and a value vector. Under a budget of $B$ tokens per layer, an eviction policy decides after each prefill chunk or every few decoded tokens which $B$ entries to keep. $\text{H}_2\text{O}$ \cite{zhang2023h2o} keeps the tokens with the largest cumulative attention received, plus a recent window; SnapKV \cite{li2024snapkv} ranks tokens by the attention of the last few prompt tokens; StreamingLLM \cite{xiao2024streamingllm} keeps the first few ``sink'' tokens and a sliding window. All of them are retrospective: they judge a token by how it has been used, or by how the current query uses it.

\textbf{Rotary positions.} Most open LLMs encode position with rotary embeddings (RoPE) \cite{su2021roformer}. A query at position $m$ and a key at position $n$ are rotated, pairwise per frequency $\theta_i$, by angles $m\theta_i$ and $n\theta_i$, so their dot product depends on positions only through the difference:
\begin{equation}
\langle R_m q, R_n k \rangle = \langle q, R_{n-m} k \rangle, \qquad R_a R_b = R_{a+b}.
\label{eq:rope}
\end{equation}
Engines cache keys after rotation. Eviction removes keys but leaves the survivors' positions untouched, so in a long session the query's position keeps growing while a kept definition stays where it was written. Two consequences follow. First, the session cannot outgrow the model's positional range even though only $B$ tokens are resident; context-extension methods such as YaRN \cite{peng2024yarn} enlarge that range by rescaling frequencies. Second, the rotary gap between the query and a kept token grows with the session.

\begin{figure}[t]
\centering
\begin{tikzpicture}[
    node distance=0.8cm and 1.1cm,
    box/.style={rectangle, rounded corners=3pt, draw=black!75, line width=0.6pt, align=center, font=\small, inner sep=6pt, minimum height=1.15cm},
    pacebox/.style={box, fill=boxblue},
    extbox/.style={box, fill=boxgray},
    arr/.style={-{Stealth[scale=0.9]}, line width=0.65pt, draw=black!80}
]
\node[extbox] (planner) {Agent planner\\dependency DAG};
\node[extbox, right=2.0cm of planner] (serving) {Serving layer\\blueprint $\rightarrow$\\token spans};

\node[pacebox, below=1.0cm of planner] (retention) {Plan-aware\\retention\\foresight\\bonus, Eq.~(\ref{eq:bonus})};
\node[pacebox, right=1.1cm of retention] (bounded) {Bounded\\positions\\renumber $\le L$,\\exact re-rotation};
\node[pacebox, right=1.1cm of bounded] (persistent) {Persistent session\\suffix-only turns};

\node[pacebox, below=1.05cm of bounded] (kvcache) {GPU KV cache\\budget $B$ tokens};
\node[extbox, right=3.0cm of kvcache] (evicted) {Evicted context\\host RAM $\rightarrow$\\disk $\rightarrow$ remote};

\draw[arr] (planner) -- node[above, font=\footnotesize] {blueprint} (serving);
\draw[arr] (serving.south) -- ++(0,-0.38) -| (retention.north);
\draw[arr] (retention) -- (bounded);
\draw[arr] (persistent) -- (bounded);
\draw[arr] (bounded) -- node[right, font=\footnotesize] {keep top-$B$} (kvcache);

\draw[arr] ([yshift=5pt]kvcache.east) -- node[above, font=\footnotesize] {evict} ([yshift=5pt]evicted.west);
\draw[arr] ([yshift=-5pt]evicted.west) -- node[below, font=\footnotesize] {restore (planned)} ([yshift=-5pt]kvcache.east);

\begin{scope}[on background layer]
\node[draw=paceblue!80!black, dashed, line width=0.8pt, rounded corners=5pt, fit=(retention) (persistent) (kvcache) (evicted), inner sep=10pt, label={[anchor=south east, font=\small\bfseries, text=paceblue!85!black]north east:PACE (KV-cache manager)}] {};
\end{scope}
\end{tikzpicture}
\caption{PACE inside the engine's KV-cache manager. The planner's dependency graph becomes a per-span foresight bonus that decides which tokens stay; bounded positions renumber the kept tokens so their rotary distance to the query stays below $L$; persistent sessions carry the cache across steps; evicted context is tiered and restored when the plan needs it.}
\label{fig:fig2}
\end{figure}
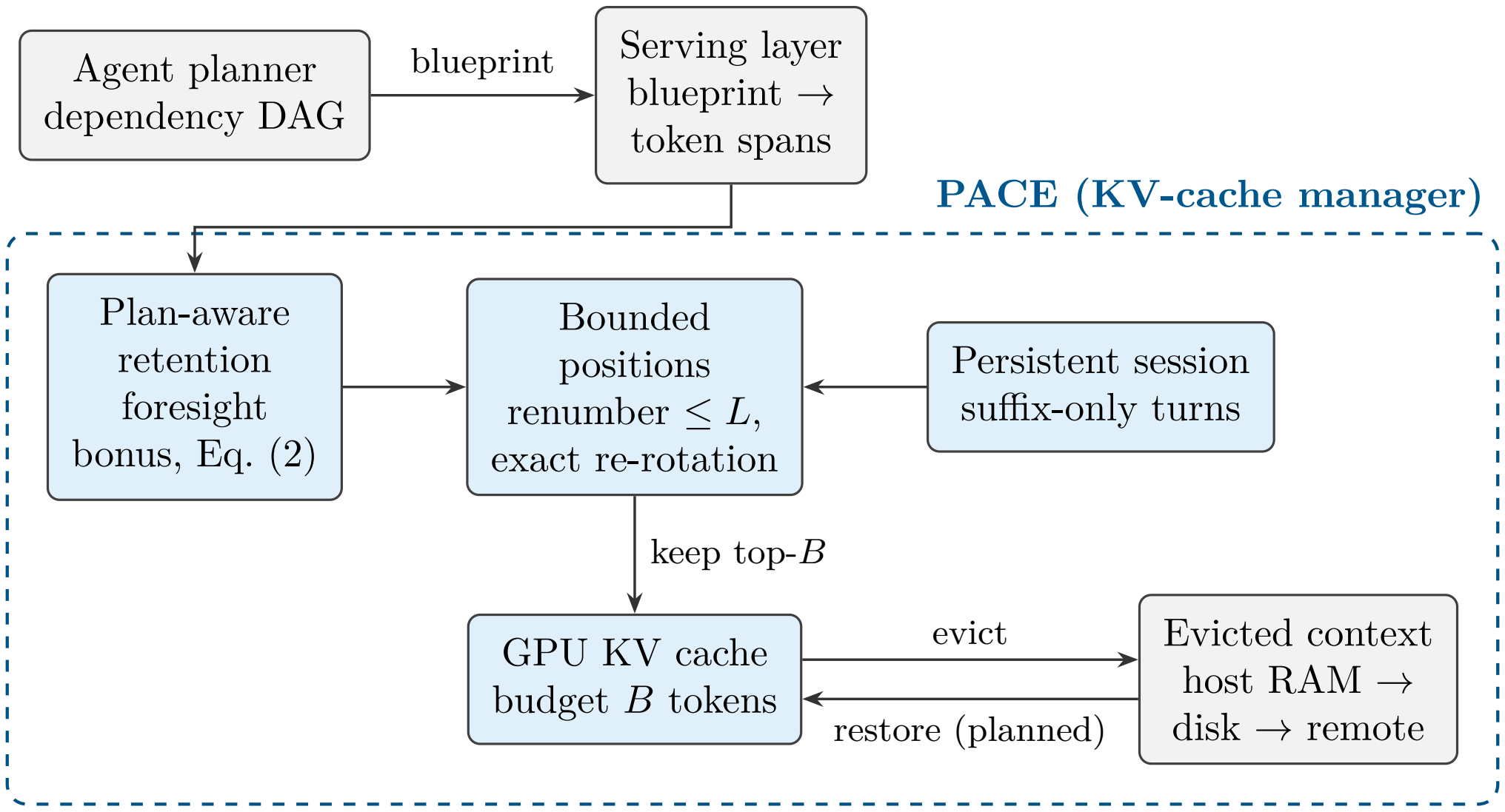

\textbf{Agent plans.} Agent frameworks commonly decompose a task into steps with explicit dependencies before executing it, for example a directed acyclic graph in which step 7 depends on steps 1 and 3. That graph is a statement about the future that the cache never sees: it says which context each upcoming step will read.

\textbf{Problem statement.} Let step $t$ issue its query at logical (session) position $m_t$, which the model sees at physical rotary position $\hat{m}_t$, and let $N_t$ be the set of earlier tokens it needs (the spans its dependencies point to). With $C_t$ the resident set, define the residency $\rho_t = |N_t \cap C_t| / |N_t|$ (minimum over layers), and the rotary gap $\Delta_t = \hat{m}_t - p(N_t)$, the distance in physical rotary positions, as the model sees them, from the query to the start of the needed span (where without renumbering, $\hat{m}_t = m_t$ and $p(N_t) = \operatorname{pos}(N_t)$). Eviction policies, including agent-aware ones, maximize $\rho_t$. This paper asks whether $\rho_t \approx 1$ suffices, and shows that it does not: at $\rho_t \ge 0.99$, accuracy still falls with $\Delta_t$ (Section~\ref{sec:mechanism}). PACE therefore controls both quantities.

\section{PACE}
\label{sec:pace}

PACE sits inside the inference engine's KV-cache manager (Figure~\ref{fig:fig2}). It changes three things: which tokens are kept (Section~\ref{sec:retention}), which positions the kept tokens carry (Section~\ref{sec:bounded}), and how a session's state lives across steps and memory tiers (Sections~\ref{sec:persistent} and \ref{sec:rehydration}). With all three disabled it reduces to $\text{H}_2\text{O}$ token for token (Section~\ref{sec:scale}), which makes every ablation a controlled comparison.

\subsection{Plan-aware retention}
\label{sec:retention}

\textbf{Blueprint.} Before each step, the agent sends the planner's dependency information as a blueprint: a list of segments, each naming earlier messages of the session and the future steps that will read them. The serving layer renders the chat template, locates each message's token span, and converts the blueprint into a forward-reference table of spans $(a, b, F)$, where $F$ is the set of future steps reading tokens $[a, b)$. No model change is required; the blueprint travels as an extra request field.

\textbf{Foresight bonus.} At step $t$, a token in span $(a, b, F)$ with next reference $s^\star = \min\{s \in F : s \ge t\}$ receives
\begin{equation}
\beta_t = \sigma \left( \frac{1}{2} + \frac{1}{2} e^{-\lambda (s^\star - t)} \right),
\label{eq:bonus}
\end{equation}
and tokens with no future reference receive $0$. The bonus is added to the token's cumulative attention score, and eviction keeps the top-$B$ by the sum. The scale $\sigma$ ($10^5$, giving a floor of $\sigma/2 = 50\text{K}$) exceeds every cumulative attention score we observe (at most 25K, in 2M-token sessions), so every referenced token outranks every unreferenced one, the core commitment of foresight. Among referenced tokens, the one needed soonest ranks highest, which is Belady's rule \cite{belady1966study} when protected spans exceed the budget; attention still breaks ties and ranks everything unreferenced. We use $\lambda = 0.35$. When a span's last planned reader has run, its bonus drops to zero and it becomes evictable. In persistent sessions (Section~\ref{sec:persistent}) the bonus of every resident token is re-computed at each turn boundary.

\subsection{Bounded positions}
\label{sec:bounded}

\textbf{Renumbering.} PACE keeps, for every resident slot, a physical position $p$, and for every sequence a shift $c$ with $\hat{m} = m - c$ for newly written tokens. Initially $c = 0$, so physical and logical positions coincide and PACE behaves exactly like $\text{H}_2\text{O}$. When the next physical position would come within a headroom $H$ of a limit $L$ (checked after each eviction), the resident row of each layer is renumbered to $p' \in \{0, \dots, R-1\}$ in logical order, every kept key is re-rotated from $p$ to its new position $p'$, and the shift is advanced so that the next token continues at $\hat{m} = R$. Physical positions therefore never exceed $L$, regardless of session length.

\textbf{Proposition 1 (Exact re-rotation).} \emph{Let a cached key be $\tilde{k} = R_p k$ for a pre-rotation key $k$ and physical position $p$. Then $R_{p'-p}\tilde{k} = R_{p'}k$: re-rotating the cached key by the position difference yields exactly the key the engine would have cached at $p'$. Consequently a query at physical position $\hat{m}$ scores the renumbered key as if it were $\hat{m} - p'$ positions away.}

The proof is Equation~(\ref{eq:rope}). We compute $R_{p'-p}$ from the engine's own cosine and sine tables per frequency $\theta_i$:
\begin{align*}
\cos((p'-p)\theta_i) &= \cos(p'\theta_i)\cos(p\theta_i) + \sin(p'\theta_i)\sin(p\theta_i), \\
\sin((p'-p)\theta_i) &= \sin(p'\theta_i)\cos(p\theta_i) - \cos(p'\theta_i)\sin(p\theta_i),
\end{align*}
normalized to unit length, so re-rotation matches the engine's rotation convention and any frequency scaling it applies; it extends to partial and interleaved rotary layouts by applying the same rotation to the rotated dimensions only.

\textbf{Semantics.} Renumbering is order-preserving and gap-collapsing: kept tokens keep their relative order, and the gaps left by evicted tokens disappear. The model therefore sees a kept definition at most $L$ positions away instead of tens of thousands. Between renumbers, ordinary relative distances apply. This is a deliberate change of what the model observes; Section~\ref{sec:scale} measures its cost.

\textbf{Choosing $L$.} The limit must leave room for the resident set and for the tokens appended between two checks: $L > B_{\max} + H$, where $B_{\max}$ is the largest resident row (the budget plus the eviction interval) and $H$ the prefill chunk. With $B = 2{,}048$ and 4,096-token chunks, $L = 8{,}192$ is the smallest admissible power of two. We show in Section~\ref{sec:mechanism} that smaller is better, so the rule is: set $L$ as small as the budget allows. Setting $L$ to the model's full positional range (40,960 for Qwen3-8B) bounds positions too, but leaves rotary gaps of ${\sim}24\text{K}$ and is measurably worse.

\textbf{Non-compounding re-rotation.} Keys are stored in $\text{bf16}$. Re-rotating the stored key at every renumber adds one rounding each time its position changes, and planner-protected keys are precisely the long-lived ones. At 100K tokens (about 23 renumbers per session with $L = 8\text{K}$) the accumulated error is small; emulating the rounding of a 1M-token session degraded accuracy from $100\%$ to $84\%$ on the same tasks, and the fix below restores $100\%$ under the same stress (Appendix~\ref{app:drift}). PACE therefore copies each resident key into a reference store at its first renumber, together with the position $p_{\text{ref}}$ it is valid at, and derives every later position from that copy in $\text{fp32}$. The error stays at about two roundings however many renumbers a key survives (in a CPU simulation of eviction churn, maximum relative key error $0.003$ after 60 position-changing renumbers, against $0.031$ when re-rotating the stored key). The store is sized to the resident set (4,096 tokens per layer, $0.3\text{\,GB}$ for Qwen3-8B); entries are invalidated when a slot is reused, moved with the slot when the engine compacts rows, and checked against the stored key before use.

\subsection{Persistent sessions}
\label{sec:persistent}

An agent typically re-sends its full transcript at every step. Without persistence, each step re-prefills that transcript; tokens evicted in earlier steps are silently re-inserted, so eviction quality never matters across steps, and the prefill cost grows quadratically with the session. PACE maps the transcript onto a persistent engine sequence: each step sends only its new messages, appended to the engine's own token history, and the KV state, including every past eviction decision, carries over. Message spans are tracked in session coordinates, so blueprints refer to the right resident tokens. For chat templates that rewrite earlier turns (for instance, ones that move the system prompt into the latest user turn), we use a prefix-stable variant of the same format so that appending is exact.

\subsection{Tiered rehydration}
\label{sec:rehydration}

Planners make mistakes, and some workloads protect more context than the budget holds. At every compaction PACE offloads the evicted rows to pinned host memory, spills to local storage under pressure, and asynchronously replicates them to a remote Redis-compatible cache (Valkey). Before a step, the planner's lookahead identifies needed spans that are no longer resident; PACE restores them from the nearest tier, splices them into the row, and renumbers so positions stay ordered and bounded. Writes to the remote tier run off the decode path.

\subsection{Implementation}
\label{sec:impl}

PACE is implemented as a subclass of the $\text{H}_2\text{O}$ cache manager in SparseEngine \cite{sparseengine2026}, an open-source inference engine, and is enabled by a configuration switch, so PACE and $\text{H}_2\text{O}$ share the same prefill, eviction schedule, and decode path. An OpenAI-compatible server accepts the blueprint as an extra request field and exposes persistent sessions. Bounded positions currently cover full-rotary, split-half RoPE with static frequency scaling (default, linear, YaRN, Llama-3 style), which includes the Qwen and Llama-architecture families; nothing in the method is specific to a model family.

\textbf{Code availability.} The experiments in this paper ran on PACE applied as an overlay to a pinned SparseEngine version. PACE's plan-aware retention and bounded positions have since been merged into SparseEngine itself \cite{sparseengine2026} (\url{https://github.com/CURRENTF/SparseEngine}, Apache License 2.0), where sessions persist through the engine's chain cache; tiered rehydration is not part of that integration. PACE is selected with \texttt{sparse\_method="pace"}, bounded positions with \texttt{pace\_bounded\_positions=True}, and a request passes its blueprint as a \texttt{PACEPlan}, a list of token spans each tagged with the future steps that will read it, alongside the prompt.

\section{Experimental Setup}
\label{sec:setup}

\subsection{A long-horizon coding benchmark}
Public long-context benchmarks mostly pose a single question over a long document \cite{hsieh2024ruler}. Agent sessions differ in two ways that matter for eviction: context arrives over many turns, and the step that needs a definition comes long after it, with unrelated work in between. We therefore built an execution-scored session benchmark in which recall of early definitions is the only path to a correct late answer.

\textbf{Session.} Each session is a sequence of user steps answered by the model under test (Appendix~\ref{app:templates} lists every template). Three definition steps fix a specification: step 1 a dataclass with randomized class and field names, two fee constants (basis points and a cap) and an exception class with a constructor code; step 2 a status class with three randomized members and values; step 3 a fee rule given as exact code, with a randomized rounding mode and quantum. Then follow filler steps, each pasting a ${\sim}2{,}000$-token chunk of Python standard-library source~\cite{python312stdlib} (PSF License) and asking for the name of its first function, so the session grows with realistic code while the answers stay short. In the 100K experiments of Sections~\ref{sec:headline} to \ref{sec:planner} (except the replication in Section~\ref{sec:headline}), each session draws its filler from a fixed pool in which files recur; later runs (the longer sessions and the drift checks) use non-repeating chunks and exclude chunks on the task's own topic (Appendix~\ref{app:integrity}). All arms of a run see identical sessions. Three probe steps ask for code that is correct only if the definitions are recalled: \texttt{fee\_for(entry)} ($p_1$), \texttt{settle(entries)} ($p_2$), and the complete module ($p_3$). Specifications are sampled per seed (30 seeds give 30 distinct specifications), so values cannot be guessed, and an automated check verifies that no probe text reveals any of them.

\begin{table}[t]
\centering
\caption{Arms. All share the KV budget ($B = 2{,}048$ tokens, including 4 sink and 128 recent tokens), the engine, greedy decoding, persistent sessions, and the eager decode path. Arms without bounded positions use $\text{YaRN}\times 4$ so that 100K-token sessions fit the positional range.}
\label{tab:tab1}
\vspace{4pt}
\small
\setlength{\tabcolsep}{4.0pt}
\begin{tabular}{@{}llll@{}}
\toprule
Arm & Retention & Positions & Role \\
\midrule
$\text{H}_2\text{O}$ & cumulative attention & as written (YaRN) & standard eviction \\
SnapKV & current-query attention & as written (YaRN) & standard eviction \\
retention only & $\text{H}_2\text{O}$ + foresight bonus & as written (YaRN) & ablation: no bounded positions \\
positions only & $\text{H}_2\text{O}$ & bounded, $L = 8\text{K}$ & ablation: no foresight \\
PACE & $\text{H}_2\text{O}$ + foresight bonus & bounded, $L = 8\text{K}$ & full system \\
PACE ($L = 40\text{K}$) & $\text{H}_2\text{O}$ + foresight bonus & bounded at the model's range & limit ablation \\
\bottomrule
\end{tabular}
\end{table}

\textbf{Scoring.} Each probe's code is executed against hidden unit tests in a subprocess: $p_1$ has 7 checks (one fee per amount), $p_2$ 3 (total, status update, duplicate detection with the right exception code), $p_3$ 5 (definitions, fees, total, status, duplicates). For $p_1$ and $p_2$ the reference classes are injected before and after the answer, so only recall of names, values and rules is tested; $p_3$ is scored standalone with a fixed standard-library import prelude identical for every arm. Independently of the tests, a recall metric checks whether the answer contains each session fact (class, field, constant names and values, rounding mode, quantum, status names and values). Together they classify each probe as pass, failed with all facts recalled (a coding error), or failed with facts missing (forgetting). A control run with no eviction (an 8K-token session within an 8K budget) scores $100\%$ on Qwen3-8B, so failures in the long sessions are attributable to the cache.

\textbf{Layouts.} In the single layout the definitions sit $20\%$ into a 100K-token session and the probes at roughly 52K, 74K and 98K tokens. In the distance layout the session length is fixed at 100K tokens and the definitions are placed $D$ tokens before the three probes at the end, so only the gap varies.

\subsection{Systems compared}
Table~\ref{tab:tab1} lists the arms. SnapKV receives the same total budget, split into sink, scored and recent tokens. The arms with foresight also run tiered rehydration; it was rarely triggered, because in these sessions the protected set fits the budget (Section~\ref{sec:scale} evaluates rehydration directly). Unless stated otherwise, the blueprint comes from the session's true dependency graph (an oracle planner); Section~\ref{sec:planner} replaces it with an LLM planner.

\textbf{Models and hardware.} Qwen3-8B \cite{yang2025qwen3} (native positional range 40,960) is the primary model. Mistral-7B-Instruct-v0.3 \cite{jiang2023mistral} (32,768; run through the engine's Llama code path, which its architecture matches exactly) is the second. All runs use one NVIDIA L4 GPU (24\,GB) with 128\,GiB of host memory.

\textbf{Statistics and integrity.} The unit of analysis is a $(\text{session}, \text{probe})$ pair, scored as the fraction of its checks passed. Comparisons between arms are paired on identical sessions and use a two-sided sign test; full-pass rates use Fisher's exact test; the two-factor interaction uses a sign-flip permutation test with bootstrap confidence intervals. The headline comparison (Section~\ref{sec:headline}), the second-model analysis, the numerical-drift verification, and the 1M- and 2M-token evaluations were pre-registered: the limit, arms, seeds, primary measures and decision rules were committed before the runs, and are reported whatever the outcome. Result files carry a fingerprint of every instruction template, and the merge tool refuses to pool runs whose wording or engine settings differ. When we changed one probe's wording after an early run, the earlier trials were kept separate rather than merged (Appendix~\ref{app:integrity}).

\section{Results}
\label{sec:results}

\subsection{Headline: pre-registered comparison on fresh tasks}
\label{sec:headline}

After the distance sweep of Section~\ref{sec:mechanism} suggested $L = 8\text{K}$, we fixed the limit, the four arms of a $2 \times 2$ design and ten new seeds in advance, and ran them once. Table~\ref{tab:tab2} reports the result. PACE passes 145 of 150 checks ($97\%$) and fully passes 28 of 30 probes; $\text{H}_2\text{O}$ passes $42\%$ and fully passes 7. Paired on identical sessions, PACE is better than $\text{H}_2\text{O}$ on 22 probes, worse on none. PACE recalls all 250 session facts and has no failure with a fact missing; its two failures are coding errors. A later replication on five further fresh sessions, with non-repeating filler that excludes chunks on the task's topic (Appendix~\ref{app:integrity}), gives the same picture: PACE $75/75$ checks ($100\%$, all 15 probes fully passed, every fact recalled) against $\text{H}_2\text{O}$'s $33/75$ ($44\%$); 13 better, 0 worse ($p = 0.0002$).

\begin{table}[t]
\centering
\caption{Pre-registered comparison (Qwen3-8B, 100K-token sessions, $B = 2{,}048$, ten fresh sessions, three probes each). Checks: hidden-test checks passed. Full: probes passing every check. Paired: $(\text{session}, \text{probe})$ pairs better/worse/tied against the reference arm, two-sided sign test. Residency: fraction of needed tokens resident at the probes.}
\label{tab:tab2}
\vspace{4pt}
\small
\begin{tabular}{lcccccl}
\toprule
Arm & Foresight & $L=8\text{K}$ & Checks & Full & Recall & Paired vs.\ $\text{H}_2\text{O}$ \\
\midrule
$\text{H}_2\text{O}$ & -- & -- & $63/150$ ($42\%$) & $7/30$ & $83\%$ & -- \\
positions only & -- & \checkmark & $94/150$ ($63\%$) & $10/30$ & $86\%$ & $14/2/14,\; p = 0.004$ \\
retention only & \checkmark & -- & $94/150$ ($63\%$) & $12/30$ & $96\%$ & $15/2/13,\; p = 0.002$ \\
PACE & \checkmark & \checkmark & $145/150$ ($97\%$) & $28/30$ & $100\%$ & $22/0/8,\; p < 10^{-6}$ \\
\midrule
PACE vs. & \multicolumn{2}{l}{Paired} & \multicolumn{2}{l}{Full-pass $p_1$} & \multicolumn{2}{l}{Full-pass $p_3$} \\
\midrule
retention only & \multicolumn{2}{l}{$17/0/13,\; p = 1.5 \times 10^{-5}$} & \multicolumn{2}{l}{$9/10$ vs.\ $1/10,\; p = 0.001$} & \multicolumn{2}{l}{$9/10$ vs.\ $1/10,\; p = 0.001$} \\
positions only & \multicolumn{2}{l}{$19/1/10,\; p = 4.0 \times 10^{-5}$} & \multicolumn{2}{l}{$9/10$ vs.\ $0/10,\; p < 0.001$} & \multicolumn{2}{l}{$9/10$ vs.\ $0/10,\; p < 0.001$} \\
\bottomrule
\end{tabular}
\end{table}

\subsection{Both components are required}
\label{sec:both}

Each component alone helps: positions only and retention only both beat $\text{H}_2\text{O}$ (Table~\ref{tab:tab2}) and tie each other ($8$ better, $6$ worse; $p = 0.79$). Neither comes close to the combination, which beats each of them on every comparison in the lower panel of Table~\ref{tab:tab2}. The two components fix different failures, which the failure taxonomy makes visible. With positions only, the needed tokens are close to the query but often gone: $\text{H}_2\text{O}$'s policy evicts the details (the quantum is missing in 20 of its 30 answers, the rounding mode in 14). With retention only, the tokens are resident (residency $1.000$ at every probe) but 30--80K positions away; 8 of its 18 failed probes contain every session fact and still apply them wrongly. Only when the tokens are both kept and close does the model use them reliably.

Is the combination more than additive? At the limit of the model's own positional range ($L = 40\text{K}$), it is: in ten earlier sessions, foresight added $5.9$ points with positions as written and $33.5$ points with bounded positions, an interaction of $+27.6$ points ($95\%$ CI $[+12.0, +42.1]$, $p = 0.0014$; Appendix~\ref{app:additional}). At $L = 8\text{K}$ both single components are stronger ($69\%$ each on the per-probe mean, vs.\ $63\%$ on pooled checks), and the interaction, $+8.3$ points (CI $[-8.0, +23.8]$, $p = 0.32$), is not significant: the additive prediction is already $89\%$, leaving at most 11 points below the ceiling. We therefore claim that both components are required, which the paired tests establish, and not that they are super-additive at the recommended setting.

\begin{table}[t]
\centering
\caption{Distance sweep (Qwen3-8B, 100K-token sessions, definitions $D$ tokens before the probes, five sessions per $D$, same seeds at every $D$). Hidden-test checks passed; the rotary gap is the mean distance, in the positions the model sees, from the query to the definitions, reconstructed from the position ledger recorded at each probe. Residency is $0.992\text{--}0.996$ for every PACE-family arm at every $D$.}
\label{tab:tab3}
\vspace{4pt}
\small
\begin{tabular}{rccccrr}
\toprule
& \multicolumn{4}{c}{Checks passed} & \multicolumn{2}{c}{Rotary gap} \\
\cmidrule(lr){2-5} \cmidrule(lr){6-7}
$D$ & $\text{H}_2\text{O}$ & retention only & PACE $L=40\text{K}$ & PACE $L=8\text{K}$ & $L=40\text{K}$ & $L=8\text{K}$ \\
\midrule
2K  & $44\%$ & $93\%$ & $100\%$ & $\mathbf{97\%}$  & 2.9K  & 2.7K \\
10K & $55\%$ & $76\%$ & $97\%$  & $\mathbf{97\%}$  & 10.7K & 5.5K \\
25K & $45\%$ & $76\%$ & $56\%$  & $\mathbf{100\%}$ & 24.4K & 5.5K \\
50K & $53\%$ & $77\%$ & $64\%$  & $\mathbf{100\%}$ & 24.7K & 5.8K \\
90K & $52\%$ & $64\%$ & $93\%$  & $\mathbf{100\%}$ & 23.9K & 6.5K \\
\bottomrule
\end{tabular}
\end{table}

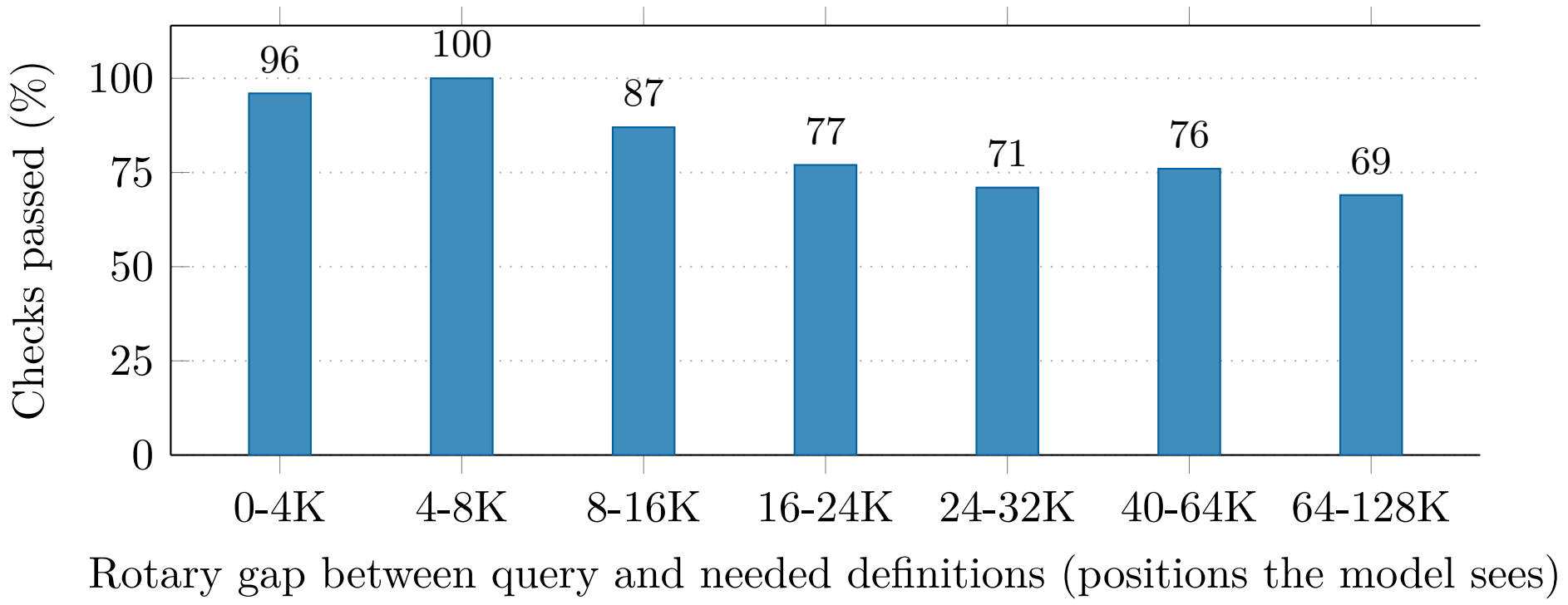
\begin{figure}[t]
\centering
\begin{tikzpicture}
\begin{axis}[
    width=12.0cm,
    height=5.0cm,
    ybar,
    bar width=14pt,
    ymin=0, ymax=114,
    symbolic x coords={0-4K, 4-8K, 8-16K, 16-24K, 24-32K, 40-64K, 64-128K},
    xtick=data,
    ytick={0, 25, 50, 75, 100},
    xlabel={Rotary gap between query and needed definitions (positions the model sees)},
    ylabel={Checks passed (\%)},
    axis y line*=left,
    ymajorgrids=true,
    grid style={dotted, gray!60},
    nodes near coords,
    every node near coord/.append style={font=\footnotesize, yshift=1pt},
    tick label style={font=\small},
    label style={font=\small}
]
\addplot[fill=paceblue!75, draw=paceblue!95!black] coordinates {
    (0-4K, 96) (4-8K, 100) (8-16K, 87) (16-24K, 77) (24-32K, 71) (40-64K, 76) (64-128K, 69)
};
\end{axis}
\end{tikzpicture}
\caption{Accuracy tracks rotary distance at full residency. All 225 probes of the three PACE-family arms in the Qwen3-8B distance sweep, binned by the reconstructed distance from the query to the definitions. Bars: checks passed. Residency of the needed tokens is 0.995--0.996 in every bin. Probes per bin, left to right: 47, 58, 30, 38, 22, 15, 15. No probe fell in 32--40K.}
\label{fig:fig3}
\end{figure}

\subsection{Mechanism: residency is not retrievability}
\label{sec:mechanism}

The distance layout fixes the session at 100K tokens and moves only the definitions (Table~\ref{tab:tab3}, Figure~\ref{fig:fig1}). $\text{H}_2\text{O}$ fails at every gap, including 2K, because a 2,048-token budget cannot hold the definitions through even one filler file; its failures are forgetting (recall $76\text{--}83\%$). Retention only keeps $99.6\%$ of the needed tokens at every gap, yet its accuracy falls from $93\%$ to $64\%$. Its recall stays at $98\text{--}100\%$, so the facts are present in the answers; what degrades is how reliably they are applied. PACE with $L = 8\text{K}$ keeps the rotary gap at 2--7K positions regardless of the logical gap and passes $97\text{--}100\%$ everywhere; it beats $\text{H}_2\text{O}$ at every gap ($p \le 0.0034$).

Figure~\ref{fig:fig3} pools every probe of every PACE-family arm by its reconstructed rotary gap. Residency is flat at $0.995\text{--}0.996$ across bins, while the pass rate falls from $96\text{--}100\%$ below 8K to $87\%$ at 8--16K and $69\text{--}77\%$ beyond 16K. Where arms overlap in a bin they score alike: the $L = 40\text{K}$ arm's packed definitions and the retention-only arm's distant ones fall in the same 16--32K bins with similar pass rates. The gap, rather than the retention policy, is what the pass rate follows. Because the largest gaps arise only at the longest $D$, this pooled view cannot fully separate gap from session position; the distance sweep, where $L = 8\text{K}$ holds the gap small at every $D$ and stays at $97\text{--}100\%$, is the controlled comparison. This is consistent with RoPE's long-range attenuation \cite{su2021roformer} and with position effects on retrieval reported for single long prompts \cite{liu2024lost}; frequency rescaling (YaRN) extends the range a model can address but, as the retention-only arm shows, does not keep distant content as usable as near content.

The $L = 40\text{K}$ arm shows why the limit should be small. Once its definitions are packed by a renumber, they sit about 24K positions from the query, and its probes in the 16--32K bins pool with retention only at the same gaps. Its spread across $D$ at a matched ${\sim}24\text{K}$ gap ($56\%$ at 25K, $64\%$ at 50K, $93\%$ at 90K) is not explained by the gap; with five sessions per point and correlated probes we report it as observed. At $L = 8\text{K}$ the spread disappears.

\subsection{Standard eviction baselines}
\label{sec:baselines}

On five further fresh sessions with the same budget, SnapKV passes 2 of 75 checks ($3\%$) against $\text{H}_2\text{O}$'s 30 ($40\%$; $11$ worse, $1$ better, $p = 0.006$). Its answers are well-formed Python with invented names, and its recall of session facts is $5\%$: none of the session's class, field or constant names survive. SnapKV ranks tokens by the attention of the current prompt's last tokens; in an agent session the current prompt is an unrelated step, so the definitions score low at the first turn after they are written and are evicted. Query-aware eviction is myopic across turns: the query that needs the definitions has not been asked yet. SnapKV was designed for compressing a single long prompt whose question comes last; we run it as the engine implements it, in multi-turn sessions with decode-time eviction. $\text{H}_2\text{O}$'s cumulative attention partly preserves the definitions but loses details: across runs it misses the quantum or the rounding mode in most of its answers.

\begin{table}[t]
\centering
\caption{Mistral-7B-Instruct-v0.3, distance layout, primary measures ($p_1 + p_2$). $D = 2\text{K}$ and $25\text{K}$: five sessions; $D = 90\text{K}$: ten (a pre-declared single extension). Residency $0.997\text{--}0.998$ for both PACE-family arms.}
\label{tab:tab4}
\vspace{4pt}
\small
\begin{tabular}{rcccccc}
\toprule
& \multicolumn{3}{c}{Checks passed ($p_1 + p_2$)} & \multicolumn{3}{c}{Recall of facts} \\
\cmidrule(lr){2-4} \cmidrule(lr){5-7}
$D$ & $\text{H}_2\text{O}$ & retention only & PACE $L=8\text{K}$ & $\text{H}_2\text{O}$ & retention only & PACE $L=8\text{K}$ \\
\midrule
2K  & $14\%$ & $72\%$ & $56\%$ & $55\%$ & $100\%$ & $100\%$ \\
25K & $0\%$  & $40\%$ & $64\%$ & $45\%$ & $91\%$  & $91\%$  \\
90K & $0\%$  & $36\%$ & $46\%$ & $67\%$ & $83\%$  & $98\%$  \\
\bottomrule
\end{tabular}
\end{table}

\subsection{With an LLM planner}
\label{sec:planner}

Foresight is only as good as the plan. We replaced the oracle with a production-style multi-agent planner: an LLM (Qwen3-8B, full context) prompted to emit a dependency graph over the session's steps, whose output is converted to blueprints by the same parser and builder an agent would use. Its edges had precision $1.0$ and recall $0.9$ in every session; the one missed edge was always from the final module to the fee rule. With these blueprints and $L = 40\text{K}$ (this experiment preceded the limit study), PACE passed 69 of 75 checks ($92\%$) against $\text{H}_2\text{O}$'s $41\%$ ($11$ better, $0$ worse, $p = 0.001$), and retention only reached $60\%$ (not significant against $\text{H}_2\text{O}$). The miss cost little because the final module also depends on the earlier \texttt{fee\_for} answer, which restates the rule, and the planner kept that edge.

\subsection{A second model}
\label{sec:mistral}

We repeated the distance study on Mistral-7B-Instruct-v0.3 under an analysis plan committed beforehand (Table~\ref{tab:tab4}). A no-eviction control first establishes the model's own ceiling: it recalls $99\%$ of facts but passes only $40\%$ of checks, failing mostly with every fact present, so for this model hidden tests largely measure coding ability. The plan therefore made $p_1 + p_2$ (scored with reference classes, insensitive to module structure) and recall the primary measures.

For a model that passes $40\%$ of checks with perfect memory, the meaningful question is how close each method comes to that ceiling. Across all probes, PACE stays at the model's own ceiling at every gap ($45\%$, $49\%$ and $37\%$ at $D = 2\text{K}$, 25K and 90K, against $40\%$ with no eviction) and recalls $96\text{--}100\%$ of facts; retention only falls below the ceiling as the gap grows ($53\%$, $32\%$, $30\%$; recall $98\%$, $94\%$, $88\%$); $\text{H}_2\text{O}$ collapses ($9\%$, $1\%$, $1\%$). The ceiling comes from a different, shorter session, so this comparison is descriptive rather than paired. PACE beats $\text{H}_2\text{O}$ significantly (at $D = 90\text{K}$, 8 better and 0 worse on $p_1+p_2$, $p = 0.008$; on all probes 12 and 0, $p < 0.001$). Most of PACE's remaining failures recall every fact and break the code, as in the control: memory is no longer the limit, the model's coding is. The pre-declared paired test of PACE against retention only at $D = 90\text{K}$ is not significant ($46\%$ vs.\ $36\%$ on $p_1+p_2$; 6 better, 4 worse; $p = 0.75$; the five added sessions, run with the non-repeating filler, favored retention only). With a $40\%$ ceiling, the two arms can differ on only a handful of probes, so the test-pass measure has little room to separate them; the separation appears in recall, which has no coding ceiling. We therefore state the second-model result as: PACE reaches the model's no-eviction ceiling and beats standard eviction on both models, while a statistically established test-pass advantage of bounded positions over retention alone is shown on Qwen3-8B.

\subsection{Exactness, cost, and scale}
\label{sec:scale}

\textbf{Exactness.} With no blueprint and no renumbering, PACE reproduces $\text{H}_2\text{O}$'s output token for token on a multi-chunk prompt, and so does PACE with bounded positions whose limit is not reached; this holds on both models. Persistent sessions produce the same tokens as re-prefilling the full history.

\textbf{Quality cost of renumbering.} Teacher-forced continuations after a 12K-token prefix, with $L = 8\text{K}$ forcing 16 renumbers, change the mean negative log-likelihood by $-0.003$ nats ($95\%$ CI $[-0.019, +0.015]$, eight windows) on Qwen3-8B and by $+0.007$ ($[-0.002, +0.016]$, four windows, 8 renumbers) on Mistral-7B. A 48K-token prompt beyond Qwen3-8B's 40,960-position range runs with physical positions below the limit and recalls a protected needle at token 0, with continuation likelihood equal to $\text{H}_2\text{O}$ on the last 30K tokens ($0.515$ vs.\ $0.516$ nats).

\textbf{Rehydration.} In a controlled test where the planner flags a definition only at the step that needs it, after it was evicted, rehydration restores it and recall is $5/5$ against $1/5$ without rehydration ($p \approx 0.05$); with no flag at all, nothing is restored and PACE equals $\text{H}_2\text{O}$. After deleting every host-side copy, all restores came from the remote tier and recall stayed $5/5$. A restore from host memory took $2.3\text{\,ms}$; one from the remote tier about $2.7\text{\,s}$.

\textbf{Time and memory.} Over ten 100K-token sessions, PACE took $1{,}047\text{\,s}$ against $\text{H}_2\text{O}$'s $963\text{\,s}$ ($+9\%$) on the same eager decode path. Persistent sessions reduce prefilled tokens $5.5\times$ and wall time $4.3\times$ against re-sending the transcript in a 16K-token session; since re-sending costs prefill quadratic in the number of steps and persistence linear, the saving grows with session length. The resident KV state is the budget (2,048 tokens, $0.30\text{\,GB}$ for Qwen3-8B) plus the reference store (at most $0.30\text{\,GB}$), independent of session length; the full cache would need $14.7\text{\,GB}$ at 100K tokens and $147\text{\,GB}$ at 1M. In a 300K-token smoke run, the latency of filler turns (a 2,000-token chunk plus a one-line answer) stayed at $0.9\text{--}1.0\text{\,s}$ from the first turn to the last; the per-step average at 1M tokens below also includes probe turns that generate up to 900 tokens. At 2M tokens filler turns do slow down, to $1.7\text{--}3.2\text{\,s}$ by the end of the session, in every arm alike (see Sessions of 2M tokens below).

\textbf{Sessions of 300K tokens.} Only bounded-position arms can run past the positional range, so at 300K we compare PACE with positions only and with a StreamingLLM-style baseline that spends the budget on sink and recent tokens with in-cache positions. On three fresh sessions with the definitions 270K tokens before the probes, PACE passes 45 of 45 checks and recalls every fact; positions only passes $58\%$ and recalls $79\%$ (it loses the rounding mode and quantum), and the StreamingLLM-style arm passes none and recalls $11\%$. PACE is better on 9 of 9 probes against the StreamingLLM-style arm ($p = 0.004$) and on 7 of 9, with 2 ties, against positions only ($p = 0.016$).

\begin{table}[t]
\centering
\caption{Quality at length (Qwen3-8B, one 24\,GB GPU, $B = 2{,}048$, filler excluding task-topic chunks, fresh sessions). Definitions are placed $90\%$ of the session before the probes. $\text{H}_2\text{O}$ and the YaRN arms cannot run past 131K positions; the StreamingLLM-style arm, at $0\%$ at 1M, was not re-run at 2M.}
\label{tab:tab5}
\vspace{4pt}
\small
\begin{tabular}{cclccc}
\toprule
Session & Sessions & Arm & Checks & Full & Recall \\
\midrule
\multirow{2}{*}{100K} & \multirow{2}{*}{5} & $\text{H}_2\text{O}$ & $44\%$ & $2/15$ & $81\%$ \\
 &  & PACE & $\mathbf{100\%}$ & $15/15$ & $100\%$ \\
\midrule
\multirow{3}{*}{300K} & \multirow{3}{*}{3} & StreamingLLM-style & $0\%$ & $0/9$ & $11\%$ \\
 &  & positions only & $58\%$ & $2/9$ & $79\%$ \\
 &  & PACE & $\mathbf{100\%}$ & $9/9$ & $100\%$ \\
\midrule
\multirow{3}{*}{1M} & \multirow{3}{*}{5} & StreamingLLM-style & $0\%$ & $0/15$ & $8\%$ \\
 &  & positions only & $33\%$ & $1/15$ & $74\%$ \\
 &  & PACE & $\mathbf{99\%}$ & $14/15$ & $100\%$ \\
\midrule
\multirow{2}{*}{2M} & \multirow{2}{*}{5} & positions only & $13\%$ & $0/15$ & $50\%$ \\
 &  & PACE & $\mathbf{100\%}$ & $15/15$ & $100\%$ \\
\bottomrule
\end{tabular}
\end{table}

\textbf{Sessions of 1M tokens.} Under the same design, with criteria committed before the run, we ran five 1M-token sessions on the single L4, each placing the definitions about 894K tokens before the probes (Table~\ref{tab:tab5}). PACE passes 74 of 75 checks ($99\%$), fully passes 14 of 15 probes, and recalls all 125 session facts; its one missed check is a coding error with every fact present. Positions only passes $33\%$ and recalls $74\%$ (losing the quantum and rounding mode), and the StreamingLLM-style arm passes none and recalls $8\%$. PACE is better than the StreamingLLM-style arm on all 15 probes ($p = 6 \times 10^{-5}$) and than positions only on at least 13, with none worse ($p < 0.001$). All three pre-registered criteria ($p_3 \ge 80\%$, recall $\ge 90\%$, significant wins over both baselines) are met. Inside the cache, the 2,050--2,170 resident tokens at each probe comprise 4 sinks, 410--1,070 needed tokens, about 470 old unprotected tokens and recent text; no unprotected token's attention score ever reached the smallest foresight bonus (maximum 14.6K against a floor of 50K), so accumulated heavy hitters never displaced planned context. After 242 renumbers per session, the definitions sat 3.8--7.1K positions from the query. A 1M-token session took about $13.6$ minutes (about $1.7\text{\,s}$ per step); the resident KV state was about $0.3\text{\,GB}$ plus at most $0.3\text{\,GB}$ of reference keys, where the full cache would need about $147\text{\,GB}$ (both computed from the model's dimensions).

\textbf{Sessions of 2M tokens.} With the same code, settings, seeds and decision rule, committed before the run, we doubled the length: five 2M-token sessions on the same L4, definitions 1.8M tokens before the probes. The StreamingLLM-style arm was dropped, having passed nothing at 1M. PACE passes all 75 checks, fully passes all 15 probes, and recalls all 125 facts. Positions only falls from $33\%$ at 1M to $13\%$ ($10/75$), fully passes no probe, and recalls $50\%$; every one of its failures has facts missing, most often the rounding mode and quantum. PACE is better on all 15 probes ($p = 6 \times 10^{-5}$; full-pass Fisher $p = 0.008$ at each probe), and all three pre-registered criteria are met again. The protection margin thus widens with length: without foresight, attention-ranked retention loses more of the definitions the longer they must survive, while the plan keeps them regardless. The cache looks as it did at 1M: 2,050--2,170 resident tokens, 410--1,090 of them needed, 460--490 old unprotected tokens, and needed-span residency at least $0.992$. After 480 renumbers per session (twice the 1M count), the definitions sat 4.4--7.1K positions from the query (mean 6.3K) against a logical gap of about 1.79M. Positions only sees the same rotary gaps (mean 6.1K), so at this length the difference between the arms is retention, not distance. No unprotected token reached the foresight floor, but the largest unprotected attention score rose from 14.6K at 1M to 25.1K, half the 50K floor (Section~\ref{sec:limitations}). The resident KV state is unchanged, where the full cache would need about $295\text{\,GB}$ (computed). Per-turn time, however, is not: the median filler turn rose from about $0.9\text{\,s}$ at the start of each session to $1.7\text{--}3.2\text{\,s}$ at the end, by the same factor in both arms, and a 2M-token session took about 31 minutes ($2.3\times$ the 1M time). Since attention, eviction and renumbering act on the fixed resident set, the growth comes from per-turn work that scales with the session elsewhere in the serving path; we have not yet isolated it.

\section{Related Work}
\label{sec:related}

\textbf{KV eviction and compression.} $\text{H}_2\text{O}$ \cite{zhang2023h2o} and Scissorhands \cite{liu2023scissorhands} keep tokens that past queries attended to; SnapKV \cite{li2024snapkv} and PyramidKV \cite{cai2024pyramidkv} score tokens by the attention of the prompt's final window, the latter with per-layer budgets; Quest \cite{tang2024quest} keeps the full cache but reads only query-relevant pages. These methods target single requests, where the question is known when compression happens. In multi-turn agent sessions it is not, which is why query-aware compression collapses in our benchmark (Section~\ref{sec:baselines}). PACE is complementary: its foresight bonus is added to $\text{H}_2\text{O}$'s score here, and could be added to any score-based policy.

\textbf{Agent-aware KV management.} A recent line of serving systems uses knowledge of agent workflows. KVFlow \cite{pan2025kvflow} abstracts the schedule as an agent step graph and evicts and prefetches prefix-cache entries by steps-to-execution; PBKV \cite{zheng2026pbkv} predicts future agent invocations for dynamic workflows; CacheScout \cite{zhang2026cachescout} learns agent transitions online; TokenCake \cite{bian2025tokencake} co-schedules offloading during function calls; Continuum \cite{li2025continuum} pins caches across tool calls with a time-to-live; WAKMan \cite{morisetty2026wakman} adapts the eviction policy to the traffic mix. These systems decide which requests' prefixes stay cached across invocations, a throughput problem. Closer to us, ActKV \cite{wang2026actkv} and MemDecay \cite{matam2026memdecay} evict within an agent's context, by contribution to action generation and by semantic region respectively. All of them optimize residency. Our finding concerns what happens once residency is achieved: retained context can still be unusable because of where it sits in position space. Plan-aware retention in PACE is one instance of this family; bounded positions are what the family lacks.

\textbf{Positions beyond the training range.} Position interpolation \cite{chen2023extending}, YaRN \cite{peng2024yarn}, LongRoPE \cite{ding2024longrope} and Self-Extend \cite{jin2024llm} change how positions map to rotations so that a model can address longer inputs. StreamingLLM \cite{xiao2024streamingllm} assigns positions within the cache rather than in the original text, so that a sink-plus-window working set can stream indefinitely, and InfLLM \cite{xiao2024infllm} retrieves blocks of distant context into a bounded window with its own position assignment; these are the closest precedents for bounded positions. PACE differs in three ways: positions are renumbered only when needed, by exact re-rotation of the cached keys, so the model sees ordinary distances between renumbers; renumbering preserves the order of planner-selected, non-contiguous context instead of a fixed sink-plus-window; and we show that the limit should be set as small as the budget allows, because accuracy depends on the resulting rotary gap. Studies of long-context use, such as ``lost in the middle'' \cite{liu2024lost} and RULER \cite{hsieh2024ruler}, document that models use distant context less reliably; we measure this at fixed residency inside a KV-evicting server and act on it.

\textbf{Offloading and memory tiers.} FlexGen \cite{sheng2023flexgen} and InfiniGen \cite{lee2024infinigen} move KV state between GPU and host; CacheGen \cite{liu2024cachegen} compresses and streams it over the network; PagedAttention \cite{kwon2023efficient} manages it in pages. PACE's rehydration tiers follow this line, but restores are scheduled by the plan rather than by a recency or attention heuristic.

\textbf{Agent memory.} Application-level memory systems such as MemGPT \cite{packer2023memgpt} decide what text to put into the prompt; PACE acts after that decision, on the KV state the prompt produces, and the two compose.

\section{Limitations}
\label{sec:limitations}

\begin{itemize}[leftmargin=1.8em, itemsep=2pt, topsep=2pt]
    \item \textbf{One task family.} The benchmark is a synthetic coding session with randomized specifications. It isolates recall of early definitions by design; other agent workloads (tool traces, multi-document research, multi-agent hand-offs) may weigh retention and distance differently.
    \item \textbf{Two models at 7--8B, one GPU type.} The statistically established test-pass advantage of bounded positions over retention alone is on Qwen3-8B; on Mistral-7B, whose coding ceiling is $40\%$, PACE reaches that ceiling and the advantage shows in recall (Section~\ref{sec:mistral}). The ceiling comparison uses a different session and is descriptive. Larger models, which may use distant context better, are untested.
    \item \textbf{Planned workloads only.} PACE's advantage comes from knowing future dependencies. In open-ended conversation without a plan it falls back to attention-based eviction with bounded positions ($63\%$ at 100K, against $42\%$ for $\text{H}_2\text{O}$ and $97\%$ with foresight); questions about arbitrary unflagged details of a long history need a full context window or a retrieval-driven restore, which we leave to future work.
    \item \textbf{Mostly an oracle planner.} One experiment uses an LLM planner (recall $0.9$). How accuracy degrades with planner errors beyond that is not yet characterized.
    \item \textbf{A fixed foresight scale bounds session length.} The bonus floor ($50\text{K}$) is a constant, while the largest cumulative attention score of an unprotected token grows with the session: 8.0--14.6K at 1M tokens, 11.5--25.1K at 2M. Extrapolated, it reaches the floor at about 5M tokens, after which accumulated heavy hitters could outrank planned context. Longer sessions need a relative bonus, scaled to the current unprotected maximum at each eviction; we have not yet evaluated it.
    \item \textbf{Gap collapse changes what the model sees.} Renumbering preserves order but not distance between kept tokens. Perplexity shows no measurable cost, but tasks that depend on how long ago something happened could be affected.
    \item \textbf{Rehydration evaluated in controlled tests.} In the 100K sessions the protected set fits the budget, so rehydration rarely fires, and it is disabled in the 300K, 1M and 2M runs; its benefit is shown in a scripted late-flag test.
    \item \textbf{Latency.} All session arms use the engine's eager decode path; we report relative overhead, not absolute serving throughput, and do not evaluate concurrent sessions. KV memory is independent of session length, but per-turn time is not: in 2M-token sessions the median filler turn grows $1.9\text{--}3.7\times$ from start to end, equally with and without foresight. The source, outside the bounded attention path, is not yet isolated.
    \item \textbf{No full-context reference at 100K.} We do not report the same model with its full cache at 100K tokens on this hardware; the no-eviction control is an 8K-token session.
    \item \textbf{Filler content matters.} Filler is Python standard-library source. In a 300K-token session, a probe asking for decimal-based fee code was derailed into reproducing the standard library's own decimal implementation, while the same session's final probe recalled every definition. Topical lookalikes of the answer test a model's resistance to copying rather than memory; our long-session runs therefore exclude filler on the task's topic, and real agent sessions offer no such filter. The 100K experiments also span two filler regimes (recurring vs.\ non-repeating files); comparisons are only ever made within one run.
    \item \textbf{Small samples per cell.} Five to ten sessions per configuration, three probes each. The headline comparison is pre-registered and strongly significant; some secondary patterns (the $L = 40\text{K}$ arm's spread across distances) are reported as observed.
\end{itemize}

\section{Conclusion}
\label{sec:conclusion}

The working assumption behind KV eviction, including its agent-aware variants, is that a token kept is a token usable. In long-horizon agent sessions it is not: at full residency, accuracy falls with the rotary distance between the needed context and the query. PACE keeps the context the plan says will be needed and keeps it close, by renumbering kept tokens so their distance to the query stays bounded. Under a budget of $2\%$ of a 100K-token session, this lifts hidden-test accuracy from $42\%$ to $97\%$ on fresh tasks, and neither half of the design achieves it alone. The effect does not fade with length: on a single 24\,GB GPU, PACE uses definitions written 900K tokens earlier correctly in $99\%$ of checks at one million tokens and definitions written 1.8M tokens earlier in all checks at two million, with a KV footprint under $0.7\text{\,GB}$ where the full cache would need $147\text{\,GB}$ and $295\text{\,GB}$, while bounded positions without foresight falls from $33\%$ to $13\%$. We expect the distinction between keeping and retrieving to matter for any system that bounds an agent's working memory, and we hope the residency-gap view and the benchmark protocol help evaluate them.

\newpage
\appendix

\section{Session templates}
\label{app:templates}

Placeholders in braces are sampled per seed. \texttt{\{k\}} is the step number.

{\footnotesize\begin{verbatim}
STEP 1: Define the core data model. Write a Python dataclass `{cls}` with fields
`{f_ref}: str` (unique reference), `{f_amt}: int` (amount in minor units), `{f_ccy}: str`
(ISO currency) and `status: str = 'new'`. Give the class two class-level constants:
`{fee_name} = {fee_bps}` (fee in basis points) and `{cap_name} = {fee_cap}` (maximum fee
in minor units). Also define an exception `{err}(Exception)` whose constructor takes an
integer `code` and stores it as `self.code`. Return one python code block.

STEP 2: Define the status values. Write a class `{status_cls}` with three string
constants: `{s_new} = '{v_new}'` is the NEW state, `{s_done} = '{v_done}'` is the SETTLED
state and `{s_void} = '{v_void}'` is the VOIDED state. Return one python code block.

STEP 3: Record the fee rule exactly; reply 'noted'. For an entry with amount A, fee rate
BPS and cap CAP (the two class constants from step 1):
  fee = (Decimal(A) * Decimal(BPS) / Decimal(10000) / Decimal({quantum}))
        .quantize(Decimal(1), rounding=decimal.{rounding}) * Decimal({quantum})
  fee = int(min(fee, Decimal(CAP)))

STEP {k}: Here is a source file from the codebase. <~2,000-token stdlib file>
Reply with one line: the name of the first function it defines.                 (filler)

STEP {k}: Write a function `fee_for(entry)` that returns the fee, as an int in minor
units, for an instance of the data class defined in step 1. Use the fee constants
defined on that class and apply the rounding rule from step 3 exactly. Return only the
function (with its imports) in one python code block.                               (p1)

STEP {k}: Write a function `settle(entries)` that takes a list of instances of the
step-1 data class. If two entries share the same unique reference field, raise the
exception class defined in step 1 with code={dup_code}. Otherwise set every entry's
status to the SETTLED-state constant of the step-2 status class and return the total,
over all entries, of each entry's amount field (from step 1) minus fee_for(entry).
Assume fee_for already exists. Return only the function in one python code block.   (p2)

STEP {k}: FINAL STEP. Write the complete module in one python code block: the step-1
dataclass and exception exactly as specified there, the step-2 status class, and the
functions fee_for(entry) and settle(entries) from the earlier steps. Reuse exactly the
names, fields, constants and rules defined earlier.                                 (p3)
\end{verbatim}}

\noindent The system prompt states that definitions made in earlier steps stay in force for the session. Sampled values: class name, field names, constant names, fee in $[17, 49]$ basis points, cap in $\{300, \dots, 900\}$, exception name and duplicate code, status class, members and values, rounding mode, quantum in $\{5, 10, 25\}$.

\section{Integrity protocol}
\label{app:integrity}

\begin{itemize}[leftmargin=1.8em, itemsep=2pt, topsep=2pt]
    \item \textbf{Hidden, execution-based scoring.} Tests are never shown to the model; answers are executed in a subprocess with a timeout. The $p_3$ import prelude (\texttt{decimal}, \texttt{dataclasses}, \texttt{typing}) is fixed and identical for all arms and models; it was not widened after a Mistral answer failed on an unimported \texttt{collections.defaultdict}.
    \item \textbf{Leak and guessability checks.} Automated tests assert that no probe contains any sampled value and that 30 seeds yield 30 distinct specifications.
    \item \textbf{Control.} With no eviction pressure, Qwen3-8B passes every check, so the hidden tests accept a model that remembers.
    \item \textbf{Wording changes.} After the first 100K run, the $p_2$ instruction was reworded because answers wrote a bare \texttt{amount} variable; trials run with the old wording are reported separately, never merged. A fingerprint of all templates is stored in every result file, and the merge tool refuses to pool runs with different fingerprints or engine settings (drift diagnostic, re-rotation mode).
    \item \textbf{Pre-registration.} The headline comparison, the Mistral analysis (including a single pre-declared extension at $D = 90\text{K}$), the drift decision rule, the fix verification, and the 1M- and 2M-token evaluations were committed before the corresponding runs, together with their decision rules, and are reported regardless of outcome.
    \item \textbf{Filler regimes.} A change to how filler chunks are built altered the sessions of runs made after it: earlier runs (most 100K results, including the headline comparison) draw each session's filler from a pool in which files recur; later runs use non-repeating chunks. We discovered this by checksumming the prompts of paired runs; every comparison reported is within one run, hence on identical sessions, and PACE with $L = 8\text{K}$ scores $97\text{--}100\%$ at 100K under both regimes. Result files now record a checksum of every session's prompts.
    \item \textbf{Topical-distractor filter.} For the 100K re-check, 300K, 1M and 2M runs we exclude filler chunks that mention \texttt{Decimal}, \texttt{decimal}, \texttt{quantize} or a \texttt{ROUND\_*} mode, a decision committed before those runs after a 300K session showed a probe copying the standard library's decimal implementation. Earlier results were not re-scored or re-run under the filter.
    \item \textbf{Design changes made after seeing data.} The 8K limit was chosen after the distance sweep; the headline run then tested it on fresh seeds. The earlier 40K-limit results are reported as measured.
\end{itemize}

\section{Additional results}
\label{app:additional}

\begin{table}[!htbp]
\centering
\caption{Earlier $2 \times 2$ at $L = 40\text{K}$ (Qwen3-8B, 100K tokens, ten sessions, single layout). The interaction is computed on per-probe fractions of checks.}
\label{tab:tab6}
\vspace{4pt}
\small
\begin{tabular}{lccc}
\toprule
Arm & Checks & Paired vs.\ $\text{H}_2\text{O}$ & Recall \\
\midrule
$\text{H}_2\text{O}$ & $74/150$ ($49\%$) & -- & $80\%$ \\
retention only & $82/150$ ($55\%$) & $10/7,\; p = 0.63$ & $95\%$ \\
positions only ($L = 40\text{K}$) & $84/150$ ($56\%$) & $12/7,\; p = 0.36$ & $82\%$ \\
PACE ($L = 40\text{K}$) & $138/150$ ($92\%$) & $19/2,\; p = 0.0002$ & $98\%$ \\
\midrule
\multicolumn{4}{l}{Interaction $+27.6$ points, $95\%$ CI $[+12.0, +42.1]$, sign-flip $p = 0.0014$ ($n = 30$)} \\
\bottomrule
\end{tabular}
\end{table}

\begin{table}[!htbp]
\centering
\caption{Recall of session facts in the distance sweep (Qwen3-8B).}
\label{tab:tab7}
\vspace{4pt}
\small
\begin{tabular}{rcccc}
\toprule
$D$ & $\text{H}_2\text{O}$ & retention only & PACE $L=40\text{K}$ & PACE $L=8\text{K}$ \\
\midrule
2K  & $81\%$ & $98\%$  & $100\%$ & $98\%$  \\
10K & $79\%$ & $98\%$  & $100\%$ & $100\%$ \\
25K & $76\%$ & $98\%$  & $88\%$  & $100\%$ \\
50K & $81\%$ & $99\%$  & $100\%$ & $100\%$ \\
90K & $83\%$ & $100\%$ & $100\%$ & $100\%$ \\
\bottomrule
\end{tabular}
\end{table}

\begin{table}[!htbp]
\centering
\caption{Mistral-7B-Instruct-v0.3, all probes ($p_3$ included; the no-eviction control passes $40\%$ of checks).}
\label{tab:tab8}
\vspace{4pt}
\small
\begin{tabular}{rcccccc}
\toprule
& \multicolumn{3}{c}{Checks passed} & \multicolumn{3}{c}{Recall} \\
\cmidrule(lr){2-4} \cmidrule(lr){5-7}
$D$ & $\text{H}_2\text{O}$ & retention only & PACE & $\text{H}_2\text{O}$ & retention only & PACE \\
\midrule
2K  & $9\%$ & $53\%$ & $45\%$ & $56\%$ & $98\%$ & $100\%$ \\
25K & $1\%$ & $32\%$ & $49\%$ & $46\%$ & $94\%$ & $96\%$  \\
90K ($n=10$) & $1\%$ & $30\%$ & $37\%$ & $70\%$ & $88\%$ & $99\%$  \\
\bottomrule
\end{tabular}
\end{table}

\section{Numerical drift of re-rotation}
\label{app:drift}

Each renumber that changes a key's position re-rotates it; computing in $\text{fp32}$ and storing in $\text{bf16}$ adds one rounding. In a CPU simulation with Qwen3-8B's rotary frequencies, the mean relative key error grows from $0.011$ after 23 renumbers (a 100K-token session at $L = 8\text{K}$) to $0.019$ after 71 (300K) and $0.037$ after 240 (1M). To test whether this matters, we added extra rotate-and-return passes per renumber, calibrated so that a 100K-token session accumulates the rounding of a 1M-token one, without changing any position. On the five sessions of the headline run where PACE scored $75/75$ checks, the stressed arm scored $63/75$ ($84\%$; 4 probes worse, none better), failing the pre-registered criterion of $90\%$; one session lost every $p_1$ check. Drift affects only the arms that renumber, so it cannot explain their advantage at 100K, where they score $97\text{--}100\%$; it does limit very long sessions. The reference store of Section~\ref{sec:bounded} removes the compounding. We verified it on the GPU under the same pre-registered criterion, on the same five sessions: with the identical stress, the reference-store implementation passes $75/75$ checks (0 worse), and without stress also $75/75$. Every resident key fit in the store (no overflow), and the consistency check never found a reference that disagreed with its stored key. All results at 300K, 1M and 2M tokens use this implementation.

\end{document}